\documentclass{article}
\usepackage{spconf,amsmath,graphicx,multirow}

\usepackage{times}

\newlength{\colwidth}
\title{The Microsoft 2016 Conversational Speech Recognition System}

\name{W. Xiong, J. Droppo, X. Huang, F. Seide, M. Seltzer,  
A. Stolcke, D. Yu and G. Zweig}
\address{Microsoft Research}

\begin{document}
\ninept

\maketitle

\begin{abstract}
We describe 
Microsoft's conversational speech recognition system,
in which we combine recent developments in neural-network-based acoustic and language modeling
to advance the state of the art on the Switchboard recognition task.
Inspired by machine learning ensemble techniques, the system uses
a range of convolutional and recurrent neural networks.
I-vector modeling and lattice-free MMI training provide significant
gains for all acoustic model architectures.
Language model rescoring with multiple forward and backward running RNNLMs,
and word posterior-based system combination provide a 20\% boost.
The best single system uses a ResNet architecture acoustic 
model with RNNLM rescoring,
and achieves a word error rate of 6.9\% on the NIST 2000 Switchboard task.
The combined system has an error rate of 6.2\%,
representing an improvement over previously reported results on this benchmark task. 
\end{abstract}


\section{Introduction}
\label{sec:intro}

Recent years have seen a rapid reduction in speech recognition error rates
as a result of careful engineering and optimization
of convolutional and recurrent neural networks. While the 
basic structures have been well known
for a long period \cite{pineda1987generalization,williams1989learning,waibel1989phoneme,lecun1995convolutional,lecun1989backpropagation,robinson1991recurrent,hochreiter1997long},
it is only recently that they have dominated the field as the best models for 
speech recognition. Surprisingly, this is the case for both acoustic
modeling \cite{sak2014long,sak2015fast,saon2015ibm,sercu2016very,bi2015very,qian2016very} 
and language modeling \cite{mikolov2010recurrent,mikolov2012context}. 
In comparison to standard feed-forward MLPs or DNNs, these acoustic models
have the ability to model a large amount of acoustic context with temporal
invariance, and in the case of convolutional models, with frequency invariance
as well.
In language modeling, recurrent models appear to improve over classical N-gram models
through the generalization ability of continuous word representations \cite{mikolov2013linguistic}.
In the meantime, ensemble learning has 
become commonly used in several neural models \cite{sutskever2014sequence,hannun2014deep,mikolov2012context},
to improve robustness by reducing bias and variance.

In this paper, we use ensembles of models extensively, as well as improvements to individual component models, to
to advance the state-of-the-art in conversational telephone speech recognition (CTS), which has been a benchmark speech recognition task since the 1990s.
The main features of this system are:
\begin{enumerate}
\item
An ensemble of two fundamental acoustic model architectures, convolutional neural nets (CNNs) and long-short-term memory nets (LSTMs), with multiple variants of each 
\item
An attention mechanism in the LACE CNN which differentially weights distant context \cite{yu2016deep}
\item
Lattice-free MMI training \cite{chen2006advances,povey2016purely}
\item
The use of i-vector based adaptation \cite{saon2013speaker} in all models 
\item
Language model (LM) rescoring with multiple, recurrent neural net LMs \cite{mikolov2010recurrent}
running in both forward and reverse direction
\item
Confusion network system combination \cite{sri-2000} coupled with search for best system subset, as necessitated by the large number of candidate systems.
\end{enumerate}

The remainder of this paper describes our system in detail.
Section \ref{sec:cnn+lstm} describes the CNN and LSTM models.
Section \ref{sec:sam} describes our implementation of i-vector adaptation. Section \ref{sec:lfmmi}
presents out lattice-free MMI training process.
Language model rescoring is a significant part of our system, and described in Section \ref{sec:rescoring}.
Experimental results are presented in Section \ref{sec:results}, followed
by a discussion of related work and conclusions.

\section{Convolutional and LSTM Neural Networks}
\label{sec:cnn+lstm}

We use three CNN variants. The first is the VGG architecture of \cite{Simonyan2014very}. Compared to the networks used previously in image recognition, this 
network uses small (3x3) filters, is deeper, and applies up to 
five convolutional layers before pooling. 
The second network is modeled on the ResNet architecture \cite{he2015deep}, 
which adds highway connections \cite{DBLP:journals/corr/SrivastavaGS15}, i.e. a linear transform of each layer's input to the layer's output
\cite{DBLP:journals/corr/SrivastavaGS15,ghahremani2016linearly}. The only difference is that we move the Batch Normalization node to the place right before each ReLU activation.

\begin{figure}[t]
\centering
\includegraphics[width=0.45\textwidth]{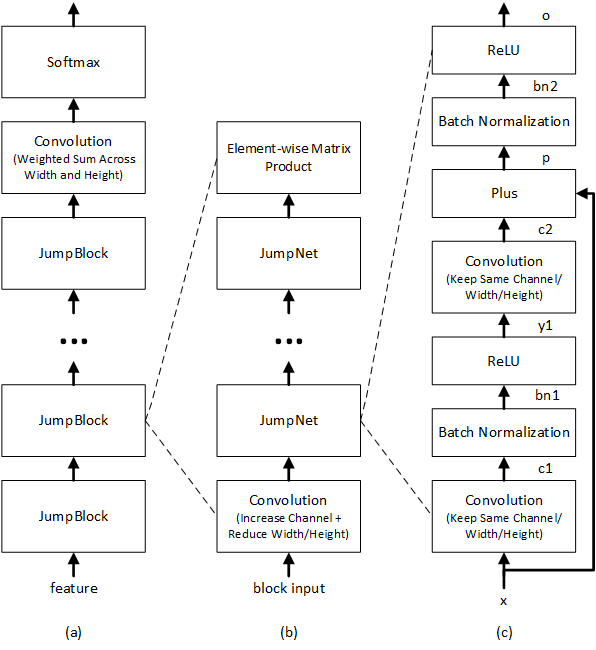}
\caption{LACE network architecture}
\label{fig:LACE}
\end{figure}

The last CNN variant is the LACE (layer-wise context expansion with attention) model \cite{yu2016deep}. LACE is a TDNN \cite{waibel1989phoneme} variant in which each higher layer is a weighted sum of nonlinear transformations of a window of lower layer frames. In other words, each higher layer exploits broader context than lower layers. Lower layers focus on extracting simple local patterns while higher layers extract complex patterns that cover broader contexts.
Since not all frames in a window carry the same importance, an attention mask is applied.
The LACE model differs from the earlier TDNN models e.g. \cite{waibel1989phoneme,waibel1989consonant} in the use of a learned attention mask and ResNet like
linear pass-through. As illustrated in detail in Figure \ref{fig:LACE}, the model is composed of 4 blocks, each with the same architecture. Each block starts with a convolution layer with stride 2 which sub-samples the input and increases the number of channels. This layer is followed by 4 RELU-convolution layers with jump links similar to those used in ResNet. 
Table \ref{tab:CNNs} compares the layer structure and parameters of the three CNN architectures.

\begin{table}[t]
\centering
\caption{Comparison of CNN architectures}
\label{tab:CNNs}
\includegraphics[width=0.45\textwidth]{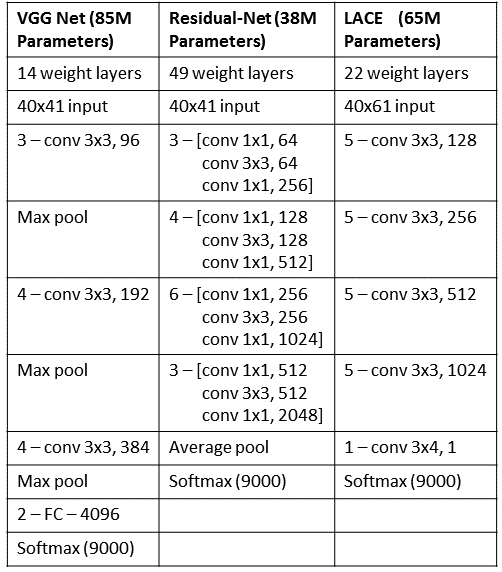}
\end{table}

While our best performing models are convolutional, the
use of long short-term memory networks is a close second. We use a 
bidirectional architecture \cite{graves2005framewise} without frame-skipping
\cite{sak2015fast}. The core model structure is the LSTM defined in 
\cite{sak2014long}. We found that using networks with more than six
layers did not improve the word error rate on the development set,
and chose 512 hidden units, per direction, per layer, as that provided a 
reasonable trade-off between training time and final model accuracy.
Network parameters for different configurations of the LSTM architecture are summarized in Table~\ref{tab:lstm}.

\begin{table}[tbh]
    \centering
\caption{Bidirectional LSTM configurations}
\label{tab:lstm}
    \small
    \begin{tabular}{|c|c|c|c|c|}
    \hline
Hidden-size & Output-size & i-vectors & Depth & Parameters  \\ \hline
512 & 9000 & N & 6 & 43.0M \\ \hline
512 & 9000 & Y & 6 & 43.4M \\ \hline
512 & 27000 & N & 6 & 61.4M \\ \hline
512 & 27000 & Y & 6 & 61.8M \\ \hline
    \end{tabular}
\end{table}

\section{Speaker Adaptive Modeling}
\label{sec:sam}

Speaker adaptive modeling in our system is based on 
conditioning the network on an i-vector  \cite{dehak2011front}  
characterization of each 
speaker \cite{saon2013speaker,saonSRK16}.
A 100-dimensional i-vector is generated for each conversation side.
For the LSTM system,
the conversation-side i-vector $v_s$ is appended to each frame of input.
For convolutional networks, this approach is inappropriate because
we do not expect to see spatially contiguous patterns in the input.
Instead, for the CNNs, we add a learnable weight matrix $W^l$ to each 
layer, and add $W^l v_s$ to the activation of the layer before the 
nonlinearity. Thus, in the CNN, the i-vector essentially serves as an 
additional bias to each layer. Note that the i-vectors are estimated
using MFCC features; by using them subsequently in systems based on
log-filterbank features, we may benefit from a form of feature 
combination.

\section{Lattice-Free Sequence Training}
\label{sec:lfmmi}

After standard cross-entropy training, we optimize the model parameters using
the maximum mutual information (MMI) objective function. Denoting a word sequence by $w$ and its 
corresponding acoustic realization by $a$, the training criterion is
\[
	\sum_{w,a \in \text{data}} \log \frac{P(w) P(a|w)} {\sum_w' P(w') P(a|w')}	\quad .
\]
As noted in \cite{sim2010sequential,vesely2013sequence},
the necessary gradient for use in backpropagation is a simple function 
of the posterior probability of a  particular acoustic model state at a given
time, as computed by summing over all possible word sequences in an 
unconstrained manner.  As first done in \cite{chen2006advances},  and more
recently in \cite{povey2016purely}, this can be accomplished with a 
straightforward alpha-beta computation over the finite state acceptor 
representing the decoding search space. In \cite{chen2006advances}, the 
search space is taken to be an acceptor representing the composition $HCLG$ for a 
unigram language model $L$ on words.
In \cite{povey2016purely}, a language model on phonemes is used instead.

In our implementation, we use a mixed-history acoustic unit language model.
In this model, the probability of transitioning into a new context-dependent phonetic state (senone)
is conditioned both the senone and phone history.
We found this model to perform better than either purely word-based or phone-based models.
Based on a set of initial experiments, we developed the following procedure:
\begin{enumerate}
\item Perform a forced alignment of the training data to select 
lexical variants and determine frame-aligned senone sequences.
\item Compress consecutive framewise occurrences of a single senone into a single occurrence. 
\item Estimate an unsmoothed, variable-length N-gram language model from this data, where the history state consists of the previous phone and previous senones within the current phone.
\end{enumerate}
\newcommand{\phone}[1]{{\it #1}}
\newcommand{\state}[2]{{\it #1}\_s#2}
To illustrate this, consider the sample senone sequence
\{\state{s}{2.1288}, \state{s}{3.1061}, \state{s}{4.1096}\},
\{\state{eh}{2.527}, \state{eh}{3.128}, \state{eh}{4.66}\},
\{\state{t}{2.729}, \state{t}{3.572}, \state{t}{4.748}\}.
When predicting the state following \state{eh}{4.66} the history consists of
(\phone{s}, \state{eh}{2.527}, \state{eh}{3.128}, \state{eh}{4.66}),
and following \state{t}{2.729}, the history is (\phone{eh}, \state{t}{2.729}).

We construct the denominator graph from this language model, and HMM transition
probabilities as determined by transition-counting in the 
senone sequences found in the training data. Our approach not only largely reduces 
the complexity of building up the language model but also provides 
very reliable training performance. 

We have found it convenient to do the full computation, without pruning, in a series of matrix-vector
operations on the GPU. The underlying acceptor is represented with a 
sparse matrix, and we maintain a dense likelihood vector for each time
frame. The alpha and beta recursions are implemented with CUSPARSE level-2
routines: sparse-matrix, dense vector multiplies. Run time is about 100 times
faster than real time.
As in \cite{povey2016purely}, we use

cross-entropy regularization. 
In all the lattice-free MMI (LFMMI) experiments mentioned below we use a trigram language model.
Most of the gain is usually obtained after processing 24 to 48 hours of data.
 
\section{LM Rescoring and System Combination}
\label{sec:rescoring}

An initial decoding is done with a WFST decoder, 
using the architecture described in \cite{mendis2016parallelizing}.
We use an N-gram language model trained and pruned with the SRILM toolkit \cite{stolcke2002srilm}.

The first-pass LM has approximately 15.9 million bigrams, trigrams, and 4grams, and a vocabulary of 30,500 words.
It gives a perplexity of 69 on the 1997 CTS evaluation transcripts.
The initial decoding produces a lattice with the pronunciation variants
marked, from which 500-best lists are generated for rescoring purposes.

Subsequent N-best rescoring uses an unpruned LM comprising 145 million N-grams.
All N-gram LMs were estimated by a maximum entropy criterion as
described in \cite{AlumaeKurimo:interspeech2012}.

\subsection{RNNLM setup}

The N-best hypotheses are then rescored using a combination of the large N-gram LM and several RNNLMs,
trained and evaluated using the CUED-RNNLM toolkit \cite{chen2016cued}.
Our RNNLM configuration has several distinctive features, as described below.

1)
	We trained both standard, forward-predicting RNNLMs and backward RNNLMs that predict words
	in reverse temporal order.
	The log probabilities from both models are added.

2)
	As is customary, the RNNLM probability estimates are interpolated at the word-level with
	corresponding N-gram LM probabilities (separately for the forward and backward models).
	In addition, we trained a second RNNLM for each direction, obtained by starting with different random 
	initial weights.
	The two RNNLMs and the N-gram LM for each direction are interpolated with weights of (0.375, 0.375, 0.25).

3)
	In order to make use of LM training data that is not fully matched to the target conversational speech domain,
	we start RNNLM training with the union of in-domain (here, CTS) and out-of-domain (e.g., Web) data.
	Upon convergence, the
	network undergoes a second training phase using the in-domain data only. Both training phases use in-domain
	validation data to regulate the learning rate schedule and termination.
	Because the size of the out-of-domain data is a multiple of the in-domain data, a standard training
	on a simple union of the data
	would not yield a well-matched model, and have poor perplexity in the target domain.

4)
	We found best results with an RNNLM configuration that had a second, non-recurrent hidden layer.
	This produced lower perplexity and word error than the standard, single-hidden-layer RNNLM architecture
	\cite{mikolov2010recurrent}.\footnote{However, adding more hidden layers produced no further gains.}
	The overall network architecture thus had two hidden layers with 1000 units each, using ReLU nonlinearities.
	Training used noise-contrastive estimation (NCE) \cite{Gutmann:NCE}.

5)
	The RNNLM output vocabulary consists of all words occurring more than once in the in-domain training set.
	While the RNNLM estimates a probability for unknown words, we take a different approach in rescoring:
 	The number of out-of-set words is recorded for each hypothesis and a penalty for them is estimated for them
	when optimizing the relative weights for all model scores (acoustic, LM, pronunciation),
	using the SRILM {\em nbest-optimize} tool.

\subsection{Training data}

The 4-gram language model for decoding was trained on the available CTS transcripts from the 
DARPA EARS program: Switchboard (3M words), BBN Switchboard-2 transcripts (850k), Fisher (21M), 
English CallHome (200k), and the University of Washington conversational Web corpus (191M).
A separate model was trained from each source and interpolated with weights optimized on RT-03 transcripts.
For the unpruned large rescoring 4-gram, an additional LM component was added, trained on 133M word of LDC
Broadcast News texts.  The N-gram LM configuration is modeled after that described in \cite{saonSRK16}, except that
maxent smoothing was used.

The RNNLMs were trained on Switchboard and Fisher transcripts as in-domain data (20M words for gradient computation,
3M for validation).
To this we added 62M words of UW Web data as out-of-domain data,
for use in the two-phase training procedure described above.  

\subsection{RNNLM performance}

Table \ref{tab:rnnlm-results} gives perplexity and word error performance for various RNNLM setups,
from simple to more complex.  The acoustic model used was the ResNet CNN.

\begin{table}
    \centering
    \caption{Performance of various versions of RNNLM rescoring.
		Perplexities (PPL) are computed on 1997 CTS eval transcripts;
		word error rates (WER) on the NIST 2000 Switchboard test set.}
\vspace*{0.1in}
	\label{tab:rnnlm-results}
    \begin{tabular}{|l|c|c|}
    \hline
	Language model					&  PPL	& WER \\
     \hline
	4-gram LM (baseline)				& 69.4	& 8.6 \\
	\  + RNNLM, CTS data only			& 62.6	& 7.6 \\
	 \quad + Web data training			& 60.9	& 7.4 \\
	 \quad \quad + 2nd hidden layer			& 59.0	& 7.4 \\
	 \quad \quad \quad + 2-RNNLM interpolation	& 57.2	& 7.3 \\
	 \quad \quad \quad \quad + backward RNNLMs	& - 	& 6.9 \\
	\hline
    \end{tabular}
\vspace*{-0.1in}
\end{table}

As can be seen, each of the measures described earlier adds incremental gains, which,
while small individually, add up to a 9\% relative error reduction over a plain RNNLM.

\subsection{System Combination}

The LM rescoring is carried out separately for each acoustic model.
The rescored N-best lists from each subsystem are then aligned into a single confusion 
network \cite{sri-2000} using the SRILM {\em nbest-rover} tool.
However, the number of potential candidate systems is too large to allow an all-out combination,
both for practical reasons and due to overfitting issues.
Instead, we perform a greedy search, starting with the single best system, and 
successively adding additional systems,
to find a small set of systems that are maximally complementary.
The RT-02 Switchboard set was used for this search procedure.
The relative weighting (for confusion-network mediated voting) of the different systems is 
optimized using an EM algorithm, using the same data, and smoothed hierarchically by 
interpolating each set of system weights with the preceding one in the search.

\section{Experimental Setup and Results}
\label{sec:results}

\subsection{Speech corpora}

We train with the commonly used English CTS (Switchboard and Fisher) corpora.
Evaluation is carried out on the NIST 2000 CTS test set, which comprises both Switchboard (SWB) and CallHome (CH)
subsets.
The Switchboard-1 portion of the NIST 2002 CTS test set was used for tuning and development.
The acoustic training data is comprised by LDC corpora 97S62, 2004S13, 2005S13,
2004S11 and 2004S09; see \cite{chen2006advances} for a full description.

\subsection{1-bit SGD Training}
All presented models are costly to train. To make training feasible, we parallelize training with our previously proposed 1-bit SGD parallelization technique \cite{seide20141}. This data-parallel method distributes minibatches over multiple worker nodes, and then aggregates the sub-gradients. 
While the necessary communication time would otherwise be prohibitive, 
the 1-bit SGD method eliminates the bottleneck by two techniques: {\em 1-bit quantization of gradients} and {\em automatic minibatch-size scaling}.

In \cite{seide20141}, we showed that gradient values can be quantized to just a single bit, if one carries over the quantization error from one minibatch to the next. Each time a sub-gradient is quantized, the quantization error is computed and remembered, and then added to the next minibatch's sub-gradient. This reduces the required bandwidth 32-fold with minimal loss in accuracy.
Secondly, automatic minibatch-size scaling progressively decreases the frequency of model updates. At regular intervals (e.g. every 72h of training data), the trainer tries larger minibatch sizes on a small subset of data and picks the largest that maintains training loss.

\vspace*{-0.1in}
\subsection{Acoustic Model Details}

Forty-dimensional log-filterbank features were extracted every 10 milliseconds,
using a 25-millisecond analysis window. The CNN models used window sizes
as indicated in Table~\ref{tab:CNNs}, and the LSTMs processed one frame
of input at a time. The bulk of our models
use three state left-to-right triphone models 
with 9000 tied states. Additionally, we have trained several models with
27k tied states. The phonetic 
inventory includes special models for 
noise, vocalized-noise, laughter and silence. We use a 30k-vocabulary
derived from the most common words in the Switchboard and Fisher corpora.
The decoder uses a statically compiled unigram graph, and dynamically
applies the language model score. The unigram graph has about 300k states and
500k arcs. All acoustic models were trained using the open-source
Computational Network Toolkit (CNTK) \cite{CNTK}.

Table \ref{tab:lfmmi} shows the result of i-vector adaptation and LFMMI training
on several
of our systems. We achieve a 5--8\% relative improvement from i-vectors, including on
CNN systems. 
The last row of Table \ref{tab:lfmmi} shows the effect of LFMMI training on the 
different models. We see a consistent 7--10\% further 
relative reduction in error
rate for all models. Considering the great increase in procedural simplicity of
LFMMI over the previous practice of writing lattices and post-processing them,
we consider LFMMI to be a significant advance in technology.

\begin{table}[tbh]
    \centering
\caption{Performance improvements from i-vector and LFMMI training on the NIST 2000 CTS set}
\label{tab:lfmmi}
	\small
    \begin{tabular}{|l|c|c|c|c|c|c|}
    \hline 
	\multirow{3}{*}{Configuration} & \multicolumn{6}{c|}{WER (\%)}                                  \\ \cline{2-7}
                               & \multicolumn{2}{c|}{ReLU-DNN} & \multicolumn{2}{c|}{BLSTM} & \multicolumn{2}{c|}{LACE}\\ \cline{2-7}
                               	& CH 	& SWB	& CH   	& SWB  	& CH  	& SWB 	\\ \hline
	Baseline          	& 21.9 	& 13.4	& 17.3 	&  10.3	& 16.9 	& 10.4  	\\ \hline
	i-vector         	& 20.1	& 11.5	& 17.6	&  9.9  & 16.4 	& 9.3     \\ \hline
	i-vector+LFMMI    	& 17.9	& 10.2	& 16.3	&  8.9	& 15.2	& 8.5		\\ \hline           
    	\end{tabular}

\end{table}

\vspace*{-0.1in}
\subsection{Comparative System Performance}

Model performance for our individual models as well as relevant comparisons from
the literature are shown in Table~\ref{tab:main}. 
Out of the 15 models built, only models given non-zero 
weight in the final system combination are shown.

\begin{table}[tbh]
    \centering
    \caption{Word error rates (\%) on the NIST 2000 CTS test set with different acoustic models
    (unless otherwise noted, models are trained on the full 2000 hours of data and have 9k senones)}
    \label{tab:main}
    \begin{tabular}{|l|l|l|l|l|}
    \hline
        \multirow{2}{*}{Model}  & \multicolumn{2}{c|} {N-gram LM} & \multicolumn{2}{c|} {Neural LM}  \\ \cline{2-5}
                                  & CH & SWB & CH & SWB \\ \hline \hline
        Saon et al. \cite{saonSRK16} LSTM        & 15.1  & 9.0  & - & - \\ \hline 
        Povey et al. \cite{povey2016purely} LSTM        & 15.3  & 8.5  & - & - \\ \hline 
        Saon et al. \cite{saonSRK16} Combination        & 13.7  & 7.6  & 12.2 & 6.6 \\ \hline  \hline
        300h ResNet         & 19.2 	& 10.0  & 17.7 & 8.2        \\ \hline
        ResNet GMM alignment    & 15.3  & 8.8  & 13.7 &  7.3       \\ \hline
        ResNet                  & 14.8 	& 8.6  & 13.2 & 6.9   \\ \hline

        VGG                  & 15.7 	& 9.1  & 14.1 & 7.6   \\ \hline
        LACE                & 14.8 	& 8.3  & 13.5 &  7.1  \\ \hline
        BLSTM         & 16.7  & 9.0  & 15.3 &  7.8 \\ \hline
        27k Senone BLSTM        & 16.2 	& 8.7  & 14.6 &  7.5 \\ \hline
        Combination             & 13.3 & 7.4 & {\bf 12.0} & {\bf 6.2}    \\ \hline
    \end{tabular}
\end{table}

\section{Relation to Prior Work}
\label{sec:prior}

Compared to earlier applications of CNNs to speech recognition
\cite{sainath2013deep,abdel2012applying}, our networks are 
much deeper, and use linear bypass connections across convolutional
layers. They are similar in spirit to those studied more recently by
\cite{sercu2016very,saon2015ibm,saonSRK16,bi2015very,qian2016very}. We improve on these 
architectures with the LACE model \cite{yu2016deep}, which iteratively
expands the effective window size, layer-by-layer, and adds an
attention mask to differentially weight distant context. Our use of 
lattice-free MMI is distinctive, and extends previous work
\cite{chen2006advances,povey2016purely} by proposing the use of a mixed
triphone/phoneme history in the language model.

On the language modeling side, we achieve a performance boost by combining multiple RNNLMs
in both forward and backward directions, and by using a two-phase training regimen to get
best results from out-of-domain data.
For our best CNN system, RNNLM rescoring yields a relative word error reduction of 20\%,
and a 16\% relative gain for the combined recognition system.
(Elsewhere we report further improvements, using LSTM-based LMs \cite{parity-techreport}.)

\section{Conclusions}
\label{sec:concl}

We have described Microsoft's conversational speech recognition system
for 2016. 
The use of CNNs in the acoustic model has proved singularly effective,
as has the use of RNN language models.
Our best single system
achieves an error rate of 6.9\% on the NIST 2000 Switchboard set. We believe
this is the best performance reported to date for a recognition system not based on system combination. 
An ensemble of acoustic models advances the state of the art to 6.2\% on the Switchboard test data.

{\bf Acknowledgments.}
We thank X. Chen from CUED for valuable assistance with the CUED-RNNLM toolkit,
and ICSI for compute and data resources.

{\footnotesize
\bibliographystyle{ieee-shortnames}
\bibliography{strings,refs}
}

\end{document}